\documentclass[conference]{IEEEtran}
\usepackage{amsmath}
\usepackage{graphicx}
\graphicspath{{./}}

\usepackage{enumitem}
\setlist[itemize]{noitemsep}

\usepackage[hang, small,labelfont=bf,up,textfont=it,up]{caption} 
\usepackage{booktabs} 

\usepackage{hyperref} 
\usepackage{psfrag}
\usepackage{pb-diagram}
\usepackage{verbatim}
\usepackage{mathtools}
\usepackage{amsmath,amsthm,amssymb,enumerate}
\usepackage{latexsym}
\usepackage{amscd}
\usepackage{amssymb}
\usepackage{mathrsfs}
\usepackage{eufrak}
\usepackage[all,arc,curve,color,frame]{xy}
\usepackage{MnSymbol}
\usepackage{float}
\usepackage{color}
\usepackage{tabu}
\usepackage{amsmath}
\usepackage{tikz}
\usepackage{qtree}
\usepackage{forest}
\usepackage{multicol}
\usepackage{enumitem}
\usepackage{yfonts}
\usepackage{subfig}
\usepackage[export]{adjustbox}
\usepackage{hyperref}

\usepackage{pdfpages}

\newlist{compactenum}{enumerate}{4}
\setlist[compactenum,1]{nolistsep}

\usetikzlibrary{automata, positioning}
\newcommand{\mb}{\mathbb}
\newcommand{\mm}{\mathrm}
\newcommand{\mc}{\mathcal}
\newcommand{\s}{\sigma}
\newcommand{\norm}[1]{\left\lVert#1\right\rVert}

\newcommand\tab[1][0.5cm]{\hspace*{#1}}

\newtheorem{theorem}{Theorem}[section]

\theoremstyle{definition}
\newtheorem{note}[theorem]{Note}

\newtheorem{defn}[theorem]{Definition}

\newtheorem{ex}[theorem]{Example}

\ifCLASSINFOpdf
\else
\fi
\begin{document}
%
\title{Applications of Graph Integration to Function Comparison and Malware Classification}

\author{\IEEEauthorblockN{Michael Slawinski}
\IEEEauthorblockA{Cylance Inc.\\
Irvine, CA\\
mslawinski@cylance.com}
\and
\IEEEauthorblockN{Andy Wortman}
\IEEEauthorblockA{Cylance Inc.\\
Irvine, CA\\
awortman@cylance.com}
}


%


\maketitle

\begin{abstract}
We classify .NET files as either benign or malicious by examining directed graphs derived from the set of functions comprising the  given file.  Each graph is viewed probabilistically as a Markov chain where each node represents a code block of the corresponding function, and by computing the PageRank vector (Perron vector with transport), a probability measure can be defined over the nodes of the given graph.  Each graph is vectorized by computing Lebesgue antiderivatives of hand-engineered functions defined on the vertex set of the given graph against the PageRank measure.  Files are subsequently vectorized by aggregating the set of vectors corresponding to the set of graphs resulting from decompiling the given file.  The result is a fast, intuitive, and easy-to-compute glass-box vectorization scheme, which can be leveraged for training a standalone classifier or to augment an existing feature space.  We refer to this vectorization technique as PageRank Measure Integration Vectorization (PMIV).  We demonstrate the efficacy of PMIV by training a vanilla random forest on 2.5 million samples of decompiled .NET, evenly split between benign and malicious, from our in-house corpus and compare this model to a baseline model which leverages a text-only feature space.  The median time needed for decompilation and scoring was 24ms. 
\footnote{Code available at \url{https://github.com/gtownrocks/grafuple}}

\end{abstract}


%
\IEEEpeerreviewmaketitle

\section{Introduction}
We classify .NET files as either malicious or benign by understanding the structural and textual differences between various types of labeled directed graphs resulting from decompilation.  The graphs under consideration are the function call graph and the set of \emph{shortsighted data flow graphs} (SDFG) derived from traversing the abstract syntax trees, one for each function in the given file. 

Each SDFG is viewed as a Markov chain and is vectorized by considering both topological features of the unlabeled graphs and the textual features of the nodes.  Under this paradigm, a heuristic notion of \emph{average file behavior} can be defined by computing expected values of specially-chosen functions $f:\mm{Vert}(G)\longrightarrow \mb{R}$ defined on the vertex sets of the given graphs against the PageRank measure.

For each graph $G$, we construct a filtration of subsets $G_{q_1}\subseteq\cdots\subseteq G_{q_{|\mathcal{P}|}}=\mm{Vert}(G)$ of $\mm{Vert}(G)$ defined by specifying a sequence of upper bounds $\mathcal{P}:=0<q_1<\dots <q_{|\mathcal{P}|}=1$  on the set of PageRank values.  The resulting sequence of expected values corresponds to a Lebesgue antiderivative $F_{f,G}$ of the function $f:\mm{Vert}(G)\longrightarrow \mb{R}$.  As there are typically many SDFG graphs per file, we vectorize by computing, for each pair $(f,q_j)$, percentiles of $\{(F_{f,G_k})_{q_j}\}_k$, where $k$ indexes the set of SDFG's present in the file.

Model interpretability is a consequence of our approach by construction, because each hand-designed function $f$, and therefore its antiderivative $F_{f,G}$, is interpretable. 

This vectorization technique and its application to malware classification are the main contributions of this paper.

\subsection{Motivation}
Static analysis classifiers trained on high dimensional data can suffer from susceptibility to adversarial examples (See \cite{Go:2019:rf} or \cite{Zh:2019:rf}) due to a large proportion of the feature space consisting of execution and semantics agnostic file features.  These include embedded unreferenced strings, certain header information, file size, etc.  See \cite{Suc:2019:rf}, \cite{Da:2017:pr}, and \cite{Liu:2018:rf} for in-depth discussions.  

Gross, K., et al. \cite{Gr:2016:rf} show that even Deep Neural Networks trained to distinguish malicious files from benign files are vulnerable to adversarial attacks.  See \cite{De:2019:pr} for a more recent example. 

Ironically, many of these features are high area under the curve features due to the copy pasta nature of most malware, but a model trained on such features can easily be tricked by perturbing these features.  This is made possible by the fact that altering these features has no effect on the runtime behavior of the file. 

Graph-based feature engineering approaches address this shortcoming by considering features extracted from the semantic structure of the file.

\subsection{Related Work}

The signature-based approach to malware detection historically has been characterized by 
hand picking features for the sake of either a rule-based approach or a regression approach as in \cite{Ra:2012:rf}.  Both signatures (hand-written static rules) and regression models fit into this category.  This approach is effective on known samples, but is prone to overfitting.  This issue was the main motivator for moving towards modeling approaches which leverage semantic structure.

The leveraging of control-flow-based vectorization of executable files for the sake of both supervised and unsupervised learning is well established in the literature, and has proven to be a technique robust to overfitting and robust to adversarial examples.  See \cite{Al:1970:rf} for one of the first such contributions.  Other early approaches involved differentiating files based on sequences of api calls.  In  \cite{Ko:2004:rf} the author builds a model based on ngrams of api calls.  See \cite{Zo:2011:rf} or \cite{Li:2018:rf} for similar approaches.

In addition to the sequential structure of function calls, one can also take into account the combinatorial graph structure of the calling relationships.  Anderson, B., et al.  \cite{An:2011:rf}  construct graph similarity kernels by viewing control flow graphs as Markov chains.  They construct a malicious/benign classifier with these kernels, which showed significant improvement over a model built only on function call ngrams. 

Chae et al. \cite{Ch:2013:rf} successfully leveraged the information present in the combinatorial structure of the control flow graph to compute the sequences and frequency of API's by considering a random walk kernels similar to those constructed above.  See \cite{Pa:2012:rf} for a similar approach.

We restrict our attention in this work to decompiled .NET, but the graph-based approach has been leveraged successfully in the similar realm of disassembly.  Indeed, \cite{Du:2005:rf} discusses the use of graph similarity to compare disassembled files, which results in a kind of file-level isomorphism useful for finding trojans.  Similar kernel methods applied to graphs arising via disassembly have been shown to be effective at detecting self-mutating malware by measuring the similarity between observed control flow graphs and known control flow graphs associated with malware.  See \cite{Bru:2006:rf}  for details.

Deep learning has also been used to extend similarity detection by constructing neural networks built on top of features derived from graph embeddings in order to measure cross-platform binary code similarity.  In \cite{Xi:2018:rf} the neural network learns a graph embedding  for the sake of measuring control flow graph similarity.  See \cite{Ph:2018:rf} for a similar approach using graph convolutional networks.  Graph embedding for the sake of measuring control flow similarity has also been applied to bug search and plagiarism detection.  See \cite{Fe:2016:rf}, \cite{Su:2014:rf}, or \cite{Sa:2015:rf} for further details and \cite{Ga:2008:rf} for a mathematical exposition of graph embedding.

Reinforcement learning has also been used in the security space to train models robust to adversarial examples created via gradient-based attacks on differentiable models, or genetic algorithm-based attacks on non-differentiable models.  See \cite{And:2018:rf} for further discussion and the authors' game-theoretic reinforcement approach to adversarial training.

Pure character-level sequence approaches (LSTM/GRU), which do not necessarily leverage the combinatorial structure of function call or control flow graphs have also been explored, as in \cite{At:2017:rf}.  The authors first train a language model in order to learn a feature representation of the file and then train a classifier on this latent representation.  See \cite{Pas:2015:rf} for a more basic RNN approach. 
 
Our approach combines a graph-based feature representation with the interpretability of a logistic regression, while avoiding the training and architectural complexity common to state-of-the-art graph convolutional neural networks.   

\section{Data}
The dataset used in this work was curated from our internal corpus and consisted of 25 million samples of .NET with 2.5 million remaining post deduplication, evenly split between benign and malicious. 

The deduplication process involved decompiling, hashing each resulting function, sorting and concatenating these hashes, and then hashing the result. 

Labels were assigned via the rule: label(file) == malicious iff any\{label$_i($file$)==$ malicious\}, where $i$ indexes the set of vendors participating on virustotal \cite{Virus:2019:rf} at the time of labeling.

\section{Decompilation of .NET}
Decompilation is a program transformation by which compiled code is transformed into a high-level human-readable form, and is used in this work to study the control flow of the files in our .NET corpus.  Program control flow is understood by studying the structure of two types of control flow graphs resulting from decompilation.  The function call graph describes the calling structure of the functions (subroutines) constituting the overall program.  The control flow of each constituent function is understood by constructing a graph from the set of possible traversals of the associated abstract syntax tree.   
\subsection{Abstract Syntax Trees}
An abstract syntax tree is a binary tree representation of the syntactic structure of the given routine in terms of operators and operands.  

For example, consider the expression $$5*3+(4+2 \,\,\%\,\, 2*8)$$ consisting of mathematical operators and numeric operands.  We may express the syntactic structure of this expression with the binary tree:

\bracketset{action character=@}
\newcount\xcount
\def\x{@@\advance\xcount1
  \edef\xtemp{$\noexpand{\the\xcount}$}%
  \expandafter\bracketResume\xtemp
}

\begin{center}
\begin{forest}
  delay={%
        content={#1}%
  },
  for tree={%
    edge path={
      \noexpand\path[->, \forestoption{edge}]
        (!u.parent anchor) -- (.child anchor)\forestoption{edge label};
    },
  }
  @+
  [+[*[5][3]][+[4][*[\%[2][2]][8]]]]
\end{forest}
\end{center}
The root and the subsequent internal nodes represent operators and the leaves represent operands.  The distilled semantic structure in this case is the familiar order of operations for arithmetic expressions. 

More generally, each node of an AST represents some construct occurring in the source code, and a directed edge connects two nodes if the code representing the target node conditionally executes immediately after the code represented by the source node.  These trees facilitate the distillation of the semantics of the program.

\subsection{Abstract Syntax Trees for the CLR}

Each node of a given AST is labeled by an operation performed on the Common Language Runtime (CLR) virtual machine.  A subset of these operations is listed as follows (see Appendix \ref{DICT} for the complete list and the details thereof):

\begin{multicols}{2}
\begin{itemize} 
\item  AddressOf
\item  Assignment
\item  BinaryOp
\item  break
\item  Call
\item  ClassRef
\item  CLRArray
\item  continue
\item  CtorCall
\item  Dereference
\item  Entrypoint
\item  FieldReference
\item  FnPtrObj
\item  LocalVar
\end{itemize} 
\end{multicols}

\begin{ex}\label{CODE_SNIPPET_II}
An AST snippet from a benign .NET sample. 
\\\\
\{\dots\\
"30":\{\\
\tab "type":"LocalVar",\\
\tab "name":"variable7"\\
\},\\
 "28":\{\\
\tab "type":"LocalVar",\\
\tab "name":"locals[0]"\\
\},\\
"29":\{\\
\tab   "type":"CLRVariableWithInitializer",\\
\tab    "varType":"System.Web.UI",\\
\tab    "name":"variable8",\\
\tab   "value":"28"\\
\},\\
"64":\{\\
\tab   "fnName":"AddParsedSubObject",\\
\tab "type":"Call",\\
\tab "target":"62",\\
\tab "arguments":[\\
\tab "63"\\
\tab]\\
\dots\}\\
\end{ex}
As shown in the example, the metadata available at each node is a function of the CLR operation being performed at that node.

\subsection{Traversals of Abstract Syntax Trees}

We consider all possible execution paths through a given abstract syntax tree and merge these paths together to form a shortsighted data flow graph (SDFG).  Consider the following code snippet:  

\begin{ex}\label{CODE_SNIPPET_I}
Small code block resulting in a nonlinear SDFG.\\

\noindent
if foo() \{\\
\tab bar();\\
\} \\
else \{\\
\tab baz();\\
\}\\
bla();\\
\end{ex} 

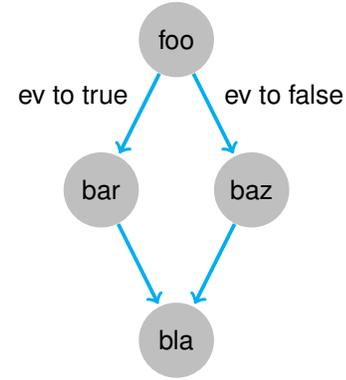
\begin{figure}[t]
\begin{tikzpicture}[scale=1][font=\sffamily]

\tikzset{node style/.style={state, 
                            minimum width=1cm,
			   draw=none,
                            fill=gray!50!white}}
                            
    \node[node style] at (1, 2)   (foo){foo};
    \node[node style] at (0, 0)         (bar){bar};
    \node[node style] at (2, 0)         (baz){baz};
    \node[node style] at (1, -2)  (bla){bla};

    \draw[every loop,
          auto=right,
          line width=0.5mm,
          =latex,
          draw=cyan,
          fill=cyan]
        (foo)     edge[auto=right]            node {ev to true} (bar)
        (foo)     edge[auto=left] node {ev to false} (baz)
        (bar)     edge[auto=right]            node {} (bla)
        (baz)     edge[auto=left] node {} (bla);
          
\end{tikzpicture}
\centering
\caption{\textbf{SDFG:}  Merging the two possible evaluation paths through this block yields the SDFG.}
\label{SDFGsimple}
\end{figure}

The two possible execution paths through this code snippet are given by $foo()\longrightarrow bar() \longrightarrow bla()$ and $foo()\longrightarrow baz()\longrightarrow bla()$.  See Figure \ref{SDFGsimple} for the resulting SDFG.

\subsection{Function Call Graphs}

The function call graph represents the calling relationships between the subroutines of the file.  The function call graphs in our corpus tended to be less linear than the SDFGs, and contained features which improved accuracy.  Notably these features were purely graph-based and were not derived via the imposition of a Markov structure, PageRank computation, or subsequent Lebesgue integration.

\section{The PageRank Vector}

The PageRank \cite{Page:1998pr} vector describes the long-run diffusion of random walks through a strongly connected directed graph.  Indeed, the probability measure over the nodes obtained via repeated multiplication of an initial distribution vector over the nodes by the associated probability transition matrix converges to the
PageRank vector, and is in practice a very efficient method for computing it to a close approximation. 

Intuitively, the PageRank vector is obtained by considering many random walks through the given graph and for each node computing the number of times we observed the walker at the given node as a proportion of all observations. See \cite{Chung:2010pr} for more details.       

Viewing the graph in question as a Markov chain we  
order the vertices $\{v_i\}$ of the graph $G$ and define the $n\times n$ probability transition matrix $T$ by
\begin{equation}
        t_{ij}=
        \begin{cases}
            1/|v_{i}^{\mathrm{out}}| & \text{if} \,\,(v_i,v_j)\in \mathrm{Edges}(G) \\
            0 &\text{otherwise}
        \end{cases}
    \end{equation}
where $v_{i}^{\mathrm{out}}$ is the set of edges emanating from vertex $v_i$ and $n=|\mathrm{Vert}(G)|$. 

\begin{figure}[t]
\includegraphics[width=5cm]{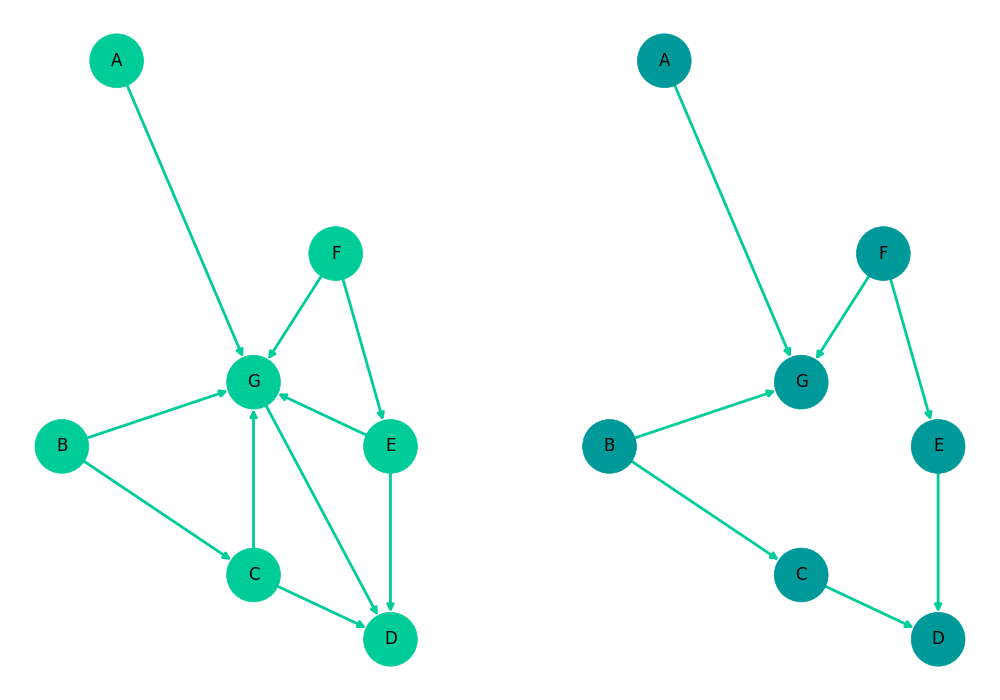}
\includegraphics[width=5cm]{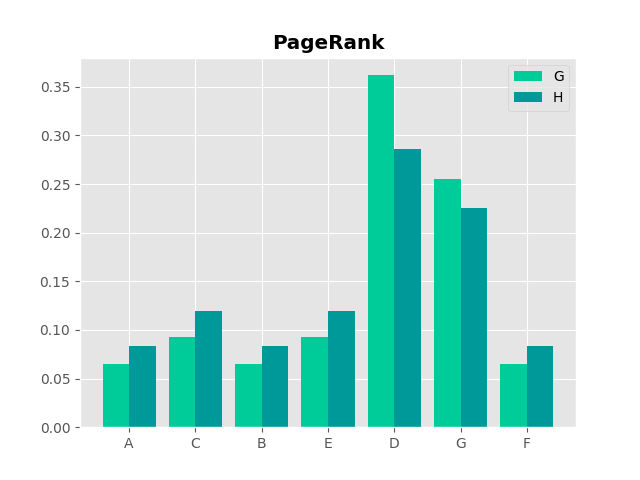}
\centering
\caption{\textbf{PageRank Distributions} The PageRank measure defined on the nodes of a given graph depends on the topology of the graph, and thus the expected values $\mb{E}[f|_{G_q}]$ of functions $f:G_q\longrightarrow\mb{R}$ also depend on the topology of the graph, where $G_q = \{v\in\mm{Vert}(G)|p_v\leq q\}$ for $p_v$ the PageRank of node $v$ and $0<q<1$.}
\label{PageRankComparison}
\end{figure}

In order to apply the Perron-Frobenius theorem, the probability transition matrix $T$ constructed via row-normalizing the adjacency matrix $A=(a_{ij})$, where $a_{ij}=1$ if there is an edge from node $i$ to node $j$ and 0 otherwise, must be irreducible.  To this end, we add a smoothing term $B$ to obtain the matrix
\begin{equation}
M = (1-p)T + pB,
\end{equation}
where 
\[
B = \frac{1}{n}\begin{bmatrix} 
    1 & 1 & \dots \\
    \vdots & \ddots & \\
    1 &        & 1 
    \end{bmatrix}
\]
The addition of the term $pB$ ensures the irreducibility of $M$ as required by the Perron-Frobenius theorem, where $p$ is the probability of the Markov chain moving between
any two vertices without traversing an edge and governs the extent to which the topology of the original graph is ignored.  See Figure \ref{PageRankComparison}. 

The resulting Markov chain is defined by 
$$P(X_t=v_i|X_{t-1}=v_j) = (1-p)t_{ij}+p\frac{1}{n},$$
where in this work we heuristically set $p=0.15$.  The sensitivity of the results to $p$ is left to a future paper.      

\begin{note}
One can view this concept in the context of a running program as the repeated calling of a particular function as represented by the SDFG, where a particular execution path is viewed as a random walk through the graph.  See \cite{Chung:2010pr} for more details.     
\end{note}

\begin{theorem}
Perron-Frobenius.  If $M$ is an irreducible matrix then $M$ has a unique eigenvector $\pi$ with eigenvalue 1.
\end{theorem}

The eigenvector $\pi$ is such that $\sum \pi_i=1$, so defines a probability measure over the vertices of $G$, which we will write as $\mb{P}_G$, or just $\mb{P}$ if the reference graph is either clear from the context or irrelevant.

\section{Integration of Functions on Graphs}

Given two labeled graphs $G,H$, and a mapping $f:\mm{Vert}(G)\sqcup\mm{Vert}(H)\longrightarrow \mb{R}$, where $f$ assigns a real number to each element $v$ of the disjoint union
$\mm{Vert}(G)\sqcup\mm{Vert}(H)$ based on the label of $v$, the connectivity at $v$, or some other scheme, a pointwise comparison of $f|_{\mm{Vert}(G)}$ and $f|_{\mm{Vert}(H)}$ may not be possible.  Consider for example the simple case of $|\mm{Vert}(G)|\neq |\mm{Vert}(H)|$.

We address this difficulty by defining a probability measure $\mb{P}_G:\mm{Vert}(G)\longrightarrow [0,1]$ for each $G\in\Gamma$, where $\Gamma$ is a set of labeled directed graph.  Then for any subset $\mathcal{I}\subset [0,1]$, we can directly compare the Lebesgue integrals $\int_{\mathcal{I}}f|_{\mm{Vert}(G)}d\mb{P}_G$ and $\int_{\mathcal{I}}f|_{\mm{Vert}(H)}d\mb{P}_H$.

Let $P$ be the PageRank vector given by the unique left eigenvector with eigenvalue 1 of the probability transition matrix of the directed graph $G$, viewed as a Markov chain.  Each file under consideration contains multiple graphs, and we wish to find a way to not only compare these graphs, but understand the ensemble of graphs in the given file.  

Let $\mathcal{P}$ be a partition of $[0,1]$ and let $G$ be a directed graph.  Let $\mb{P}_G$ be the probability measure on $\mm{Vert}(G)$ given by the PageRank vector $P=\,\langle p_v\rangle$.  Consider a function $f:\mm{Vert}(G)\longrightarrow \mb{R}$.  The function 

\begin{equation}\label{GAntiDerivative}
F_{f,G}:\mathcal{P}\longrightarrow \mb{R}
\end{equation}
\begin{equation}
q\mapsto \mb{E}[f|_{G_q}],  \nonumber
\end{equation}
where $G_q = \{v\in\mm{Vert}(G)\,|\,p_v\leq q\}$, and $\mb{E}[f|_{G_q}]=\int_{G_q}fd\mb{P}_G=\sum_{v\in G_q} f(v)p_v$.  Mathematically $F_{f,G}$ is the Lebesgue antiderivative of $f$ over $\mm{Vert}(G)$ with measure given by $\mb{P}_G$. 

The above process of building a function $F$ on $\mc{P}$ from a graph $G$ and a rule $f$ which can be applied consistently to any element of $\Gamma$ can be formulated as a mapping 
$$\Gamma\times \mm{Fun}(\bigsqcup_{\Gamma}\mm{Vert}(G),\mb{R})\longrightarrow \mm{Fun}(\mathcal{P},\mb{R})$$
\begin{equation}\label{Graph2Vec}
(G,f)\mapsto (F_{f,G}:q\mapsto \mb{E}[f|_{G_q}]),
\end{equation}
where $\mm{Fun}(X,Y)$ is the set of functions from $X$ to $Y$.

\section{Similarity Measure on Graph Space}\label{MGS}

Let $\Gamma_{\textgoth{A}}$ be the set of directed graphs with vertices labeled from the alphabet $\textgoth{A}$ and let $S$ be a set of functions defined on $G$.
Define the vectorization map 

\begin{equation}\label{VECTMAP}
\mb{V}:\Gamma_{\textgoth{A}}\longrightarrow \mb{R}^{|S||\mc{P}|}
\end{equation}
$$G\mapsto (\mb{E}[f_i|_{G_{q_j}}])$$
where the expected value is taken with respect to the PageRank measure as defined in the previous section. 

We construct a similarity function via \eqref{VECTMAP}
$$\textgoth{S}:\Gamma_{\textgoth{A}}\times\Gamma_{\textgoth{A}}\longrightarrow \mb{R}$$
\begin{eqnarray*}
    (G,H) & \mapsto  & \norm{\mb{V}(G) - \mb{V}(H)}_p \\ 
    & = &  \left(\sum_{ij}  \left|\mb{E}[f_i|_{G_{q_j}}] - \mb{E}[f_i|_{H_{q_j}}]\right|^p\right)^{\frac{1}{p}}
\end{eqnarray*}
for $f_i\in S$, $q_j\in\mathcal{P}$, and $p>=1$.

\begin{defn} A metric on a set $X$ is a function 
$$d:X\times X\longrightarrow [0,\infty)$$
satisfying
\vspace{1mm}
\begin{compactenum}[label=(\roman*)]    
\item $d(x,y)=0 \iff x=y$
\item $d(x,y)=d(y,x)$   
\item $d(x,z)\leq d(x,y)+d(y,z)$
\end{compactenum} 
\end{defn}

Condition (ii) is satisfied since $|a-b|=|b-a|$ for all $a,b\in\mb{R}$ and (iii) is satisfied since $\mb{V}(G)\in\mb{R}^k$ for all $G$ by construction. 

 However, it is possible that $\textgoth{S}(G,H)=0$ for $G\nsimeq H$, meaning that while $\textgoth{S}$ is effective as a measure of similarity of labeled directed graphs, it is not a metric on $\Gamma_{\textgoth{A}}$.   

Indeed, let $G = \{a:b\}$, $H=\{a:c\}$, and let $S=\{f\}$ where $f:\mm{Vert}(G)\longrightarrow\mb{R}$ is defined by $v\mapsto \mm{int}(\mm{label}(v)==a)$.  Then $\mb{V}(G)=\mb{V}(H)$, which implies $\textgoth{S}(G,H)=0$.  The graphs $G$ and $H$ have the same topology and the same combinatorial structure, but the set of functions $S$ is insufficient to distinguish $\mb{V}(G)$ from $\mb{V}(H)$.

Additional conditions must be imposed on $\Gamma_{\textgoth{A}}$ and the functions defined thereon in order to guarantee the injectivity of $\mb{V}$, a necessary condition for $\textgoth{S}$ to define a metric.  We leave this analysis to a future paper.

\section{Application of Lebesgue Integration on Graphs to SDFGs}

The machinery developed in the previous sections lends itself to two immediate applications.

The first is the use of the vectorization map 

\begin{equation}\label{VECT}
\mathrm{Vect}:\mm{Files}\longrightarrow \mb{R}^k
\end{equation}
\begin{equation}\label{VectSequence}
\mm{file}\xrightarrow{\mathrm{decomp}}\{G_{\mm{SDFG}}\}\xrightarrow{Eq\,\eqref{Graph2Vec}} \{v_{G_{\mm{SDFG}},f}\}
\end{equation}
applied to .NET files, constructed via decompilation followed by integration of selected functions on SDFGs as described in Equation \eqref{Graph2Vec}, to i) construct an $N$-class classifier on a given corpus of labeled .NET files, and ii) cluster these files in $\mb{R}^k$ using any of the classic metrics defined on Euclidean space. 

The second application is classification and clustering of .NET files within the metric space $\Gamma$, described in Section \ref{MGS}.  The remainder of this paper concerns the applications of the vectorization map $\mathrm{Vect}:\mathrm{Files}\longrightarrow \mb{R}^k$.

\subsection{Feature Hashing}
Feature hashing allows for the vectorization of data which is both categorical in nature and is such that the full set of categories is unknown at the time of vectorization.  We construct a hash map on strings by wrapping the hash function from the Python standard library as follows:  
$$\eta:\text{string} \mapsto \log_{10}(\max(1,|\text{hash}(\text{string})|)$$
We take a log in order to bring the integer resulting from the hash function down to a more aesthetic size.  This has no effect on the model as random forests are agnostic to the magnitudes of feature values.

\subsection{Functions on SDFGs}
For the sake of clarity, we illustrate the typical form such a function takes with an example.  Consider the SDFG snippet given in Example \ref{CODE_SNIPPET_II}.  We define the function
$$\text{CLRVariable}:v\mapsto \eta(\mm{varType}(v))$$
by 
\begin{eqnarray*}
v_{29} &\mapsto& \eta(\mm{varType}(v_{29})) \\
&=&  \eta(\text{`System.Web.UI'}) \\
&=& \log_{10}(\max(1,|\text{hash}(\text{`System.Web.UI'})|)
\end{eqnarray*}

The complete set of functions $$f:\mm{Vert}(\mm{SDFG})\longrightarrow \mb{R}$$ leveraged in this work is listed in Appendix \ref{FUNCSONSDFGS}.

\subsection{Lebesgue Integration of Functions on SDFGs}
Because the nodeset of any SDFG is finite and the PageRank measure defined thereon is discrete, the Lebesgue antiderivatives of the functions defined in the previous section take the form of sequences of dot products. 

We illustrate the nature of the map $\{G_{\mm{SDFG}}\}\xrightarrow{Eq\,\eqref{Graph2Vec}} \{v_{G_{\mm{SDFG}},f}\}$ via an example. 

\begin{ex}

Consider a SDFG G representing the traversals of some function's abstract syntax tree.  Assume $\mathrm{Vert}(G)=\{v_1,v_2,v_3,v_4\}$ and that 
PageRank$(G)=\langle p_{v_1}=0.1,p_{v_2}=0.15,p_{v_3}=0.25,p_{v_4}=0.5\rangle$.  Assume the nodes $v_1,v_4\in\mathrm{Vert}(G)$ both correspond to function calls $\phi_{v_i}(args_{v_i})$, where $args_{v_i}$ represent the set of arguments passed to $\phi_{v_i}$.    
\begin{equation}\mathrm{NumPass2Call}:\mathrm{Vert}(G)\longrightarrow\mb{R}
\end{equation}
\begin{equation}
        v_i\mapsto
        \begin{cases}
            \#\text{args}_{v_i} & if \,\,i\in\{1,4\} \\
            0 &\text{otherwise}
        \end{cases}
    \end{equation}
Take the partition of $[0,1]$ defined by $$\mathcal{P}=[0.05,0.12,0.20,0.95]$$  The Lebesgue antiderivative of NumPass2Call on $G$
$$F_{\mathrm{NumPass2Call},G}:\mathcal{P}\longrightarrow \mb{R}$$
$$q\mapsto \mb{E}[\mathrm{NumPass2Call}|_{G_q}],$$
takes the form 

$$
\begin{pmatrix}
  0.05\\
  0.12\\
  0.20\\
  0.95
\end{pmatrix}
\mapsto
\begin{pmatrix}
  0\\
  0.1*\#\text{args}_{v_1}\\
  0\\
  0.1*\#\text{args}_{v_1}+0.5*\#\text{args}_{v_4}
\end{pmatrix}
$$
\end{ex}

In general, each entry of the vector $F_{f,G} = (\mb{E}[f|_{G_{q_1}}],\dots,\mb{E}[f|_{G_{q_{|\mathcal{P}|}}}])$ is a linear combination of the form $S=\sum p_ir_i$, where $p_i$ is the element of the PageRank vector assigned to node $i$ and $r_i$ is a real number resulting from applying $f$ to node $i$.

\begin{figure}[t]
\includegraphics[width=5cm]{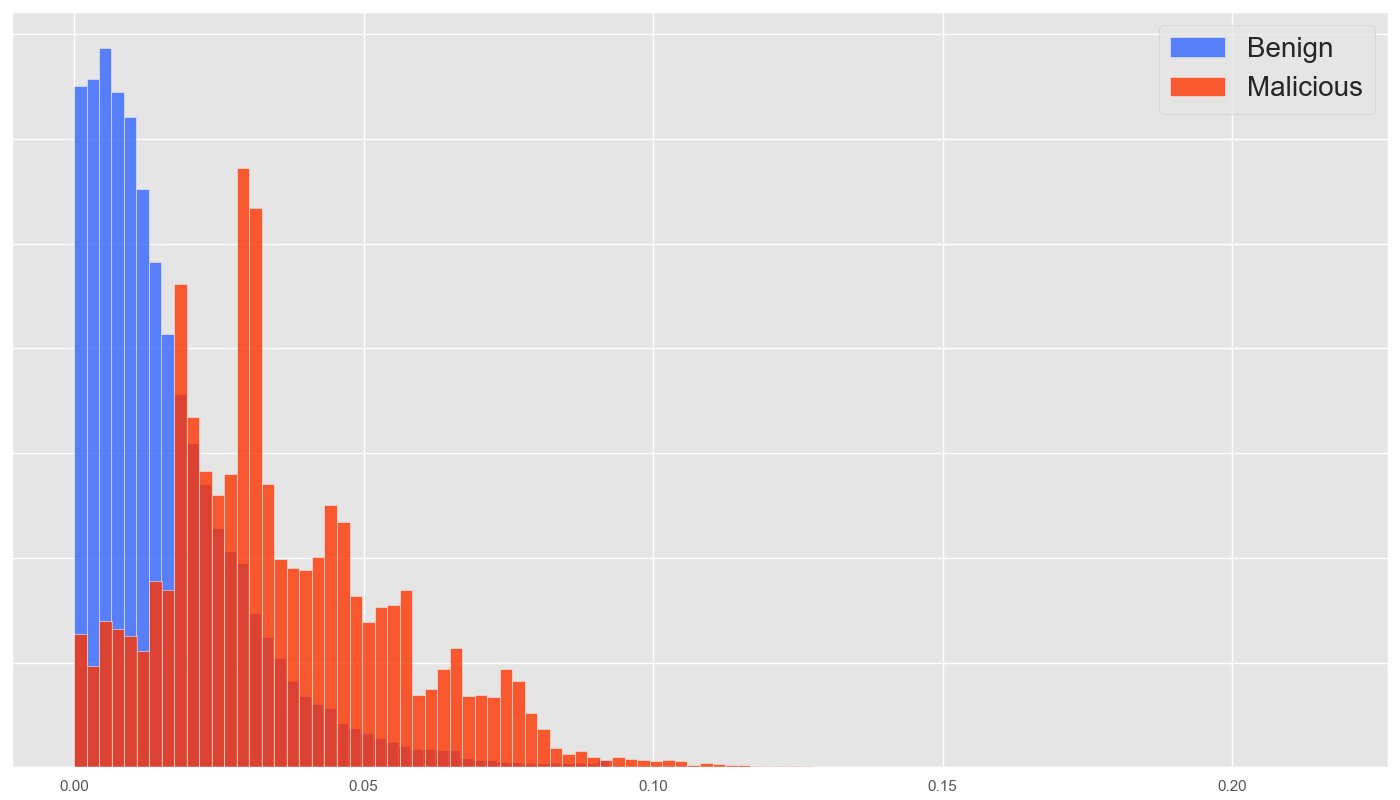}
\centering
\caption{\textbf{ClassRef\_ExpectedType\_60:} We compute the standard deviation of the set of expected values $\mb{E}[(v\mapsto (type(v)==\mm{ClassRef}))|_{G_{0.60}}]$ indexed by the set of SDFGs $G$ resulting from decompilation, where we take $type(v)==\mm{ClassRef}$ to be one of true and 0 if false.  There is one such expected value per SDFG graph and this feature is obtained by computing the mean of the set of these expected values across all SDFGs in the file.}
\label{ClassRef_ExpectedType_60}
\end{figure}


%

\section{Vectorization of Function Call Graphs}\label{VFCG}

The features of the form $f:\mm{Vert}(G)\longrightarrow \mb{R}$ extracted from the function call graphs are limited to:
\begin{itemize}
\item $\mm{Crypto Flag}$\begin{equation}
        v\mapsto
        \begin{cases}
            1 & \text{if crypto flag}\in v \\ \nonumber
            0 & \text{if crypto flag}\notin v
        \end{cases}
    \end{equation}
\end{itemize}
This function, unlike those applied to the SDFGs, is not integrated.  We simply include the cryto flag as a feature directly.

The remaining features extracted from the function call graphs are combinatorial and topological in nature.  

Let $G_C$ be the function call graph for a single .NET file and let $C=\{c_1,\dots,c_N\}$ be the connected components thereof.  Let $\deg(v)$ represent the number of edges connected to the vertex $v\in \mm{Vert}(G_C)$.  Let $L=\{|c_1|,|c_2|,\dots,|c_N|\}$ where $|c_i|$ is the number of nodes of component $i$.  We extract the following features:   
\begin{itemize}
\item max($L$)/min($L$) 
\item $N$
\item mean$(\{\deg(v)|v\in\mm{Vert}(G_C)\})$
\item std$(\{\deg(v)|v\in\mm{Vert}(G_C)\})$
\item $|\mm{Vert}(G_C)|$ 
\item $|\mm{Edges}(G_C)|$
\end{itemize}


\section{Experiments}

 We compare PMIV to a baseline method we call Uniform Measure Integration Vectorization (UMIV). 

Uniform Measure Integration Vectorization is similar to PMIV in that the method is defined by computing a graph-based integral of functions defined over the node sets, where these functions are exactly those used for PMIV.  The critical difference is that UMIV is defined via integration against the uniform measure.   

This means that instead of computing $$F_{f,G}=(\mb{E}[f|_{G_{q_1}}],\dots,\mb{E}[f|_{G_{q_{|\mc{P}|}}}])$$ 
as in defined via Equation \eqref{GAntiDerivative}, we compute
\begin{equation}\label{UniformGraphIntegration}
F_{f,G}^U:=\frac{1}{|\mm{Vert}(G)|}\sum_{v\in\mm{Vert}(G)} f(v),
\end{equation}
 i.e., a simple average of the given function over the node set of the given graph.  PMIV and UMIV similarly leverage the textual information embedded in SDFGs, but UMIV ignores the combinatorial structure of the SDFGs.

\subsection{Parsing (same for PMIV and UMIV)}
Each .NET file is decompiled resulting in i) an abstract syntax tree for each function within the file and ii) the function call graph.  The abstract syntax trees are traversed individually resulting in a single SDFG for each function within the file.  The function call graph is a directed graph indicating which functions call which other functions. 

\begin{ex}

Consider the C\# program\\

\noindent
using System;\\
\noindent
class Hello\\
\{ \\
\tab static void Main()\\
\tab \{ \\
\tab \tab Console.WriteLine("Hello, World!");\\
\tab		\}\\
	\}
\end{ex}

\noindent
Three graphs result from decompilation - an empty function call graph and two linear SDFG graphs.  See Appendix \ref{DECOMPILATION} for the decompiler output.  

\subsection{Vectorization (PMIV)}

Each file is vectorized by applying both the vectorization map \eqref{VECT} to the set of shortsighted data flow graphs (many per file) and the vectorization of the function call graph (one per file) as described in Section \ref{VFCG}. 

\subsubsection{SDFG}

Given a file marked by its hash $h$, we consider a set of SDFG graphs $\{G^h_i\}$ obtained by decompiling $h$.  

For each function $f:\sqcup_{\Gamma}\mm{Vert}(G)\longrightarrow \mb{R}$, hash $h$, and partition $\mc{P}$, we can compute the values 
\[
\left\{
\begin{array}{ccc}
F_{f,G_1}(q_1)  &  \dots &  F_{f,G_1}(q_{|\mc{P}|}) \\
  & \vdots  &  \\
F_{f,G_n}(q_1)  &  \dots &  F_{f,G_n}(q_{|\mc{P}|}) 
\end{array}
\right\}
\]
We can then compute both the mean and standard deviation of the set $\{F_{f,G^h_i}(q_j)\}$ for each $q_j\in\mc{P}$.  As the number of SDFGs varies by file, this is necessary to guarantee that every file in the corpus can be mapped to $\mb{R}^k$ for some fixed $k$.

The file $h$ is mapped, via integrating $f$ over $G$, to the feature space given by coordinates $\{\mu(\{F_{f,G^h_i}(q_j)\}_i)\}_{j}\bigcup\{\s(\{F_{f,G^h_i}(q_j)\}_i)\}_{j}$.

The file $h$ is then described by the feature vector given by
$$\bigoplus_f(\mu(\{F_{f,G^h_i}(q_j)\}_i))_j\oplus (\s(\{F_{f,G^h_i}(q_j)\}_i))_j$$
where $f$ is an index running over the set of functions $f:\mm{Vert}(G)\longrightarrow \mb{R}$. 
See Figure \ref{ClassRef_ExpectedType_60} for a resulting feature histogram for a particular $f$. 


\subsubsection{Function Call Graphs}
The function call graph features are included as components of the final file-level vector directly without computing means and standard deviations, as there is a single such graph per file.

\subsection{Vectorization (UMIV)}

Files are vectorized in essentially the same way as in PMIV, except that we assign the probability
$1/|\mm{Vert}(G)|$ to each vertex $v\in\mm{Vert}(G)$ for $G$ a SDFG.

The vectorization scheme is defined in Equation \eqref{UniformGraphIntegration}, and the final reduction of these values across SDFGs into a single vector corresponding to a single file is identical to that of PMIV.   

\subsection{Algorithm}
We train a separate random forest (\cite{Ho:1995:rf}\cite{Ho:1998:rf}) for each vectorization method, each with identical hyperparameters. 

A random forest is an ensemble learning method for classification, regression, and other tasks that operates by constructing a multitude of decision trees at training time and scoring via a polling (classification) or averaging (regression) procedure over its constituent trees. 

This algorithm is especially valuable in malware classification as scoring inaccuracy caused by unavoidable label noise is somewhat mitigated 
by the ensemble. 

\subsection{Training and Validation (same for PMIV and UMIV)}
The .NET corpus was first deduplicated via decompilation by first decompiling each file, hashing each resulting graph, lexicographically sorting and concatenating these hashes, and then hashing the result. 

The deduplicated corpus was split into training (70\%), validation (10\%), and test (20\%) sets.  We used the grid search functionality of scikit-learn with cross-validation for hyperparameter tuning of the random forest.  The optimal model is described in Table \ref{randomforesthyperparameters}.

\begin{table}[htp]
\caption{Random Forest Hyperparameters}
\begin{center}
\begin{tabular}{c|c}
max leaf nodes & None \\
\hline
min samples leaf &  1\\
\hline
warm start & False\\
\hline
min weight fraction leaf &  0\\
\hline
oob score & False\\
\hline
min samples split &  2\\
\hline
criterion &  gini\\
\hline
class weight &  None\\
\hline
min impurity split &  2.09876756095e-05\\
\hline
n estimators &  480\\
\hline
max depth &  None\\
\hline
bootstrap &  True\\
\hline
max features & sqrt
\end{tabular}
\end{center}
\label{randomforesthyperparameters}
\end{table}%

\section{Experimental Results}

\subsection{Accuracy, Precision, Recall (PMIV and UMIV)}

The model is 98.3\% accurate on the test set using only 400 features, which is tiny for a static classifier. 

The precision on malicious files was 98.94\%, meaning that of the files classified as malicious by the model, 98.94\% of them were actually malicious.  Precision on benign files was 97.88\% and recall on benign files was 99.37\%.    

The recall on malicious files was 96.47\%, meaning that of the malicious files, 96.47\% of them were correctly scored as malicious.  Of the four precision/recall values, malicious recall was the weakest.  There are very likely features of malicious .NET files that are not captured by the set of functions $f:\mm{Vert}(G)\longrightarrow \mb{R}$ we currently leverage to construct our feature space.      

As shown in tables \ref{PMIV_TABLE} and \ref{UMIV_TABLE}, our graph-structure-based vectorization method PMIV outperforms our baseline UMIV method by wide margins, demonstrating the efficacy of our graph integration construction.


\begin{table}
\caption{PMIV Performance}\label{PMIV_TABLE}
\resizebox{\columnwidth}{!}{
\begin{tabular}{llllr}
\cmidrule(r){1-5}
Class & Precision & Recall & F1-score & Support\\
\midrule
Benign & 97.88\% & 99.37\% & 98.62\% & 696827\\
Malware & 98.94\% & 96.47\% & 97.69\% & 424420 \\
\midrule
avg/total & 98.28\% & 98.27\% & 98.27\% & 1121247 \\
\midrule
False Positive Rate & 1.10\%\\
False Negative Rate & 1.72\%\\
\bottomrule
\end{tabular}
}
\end{table}

\begin{table}
\caption{UMIV Performance}\label{UMIV_TABLE}
\resizebox{\columnwidth}{!}{
\begin{tabular}{llllr}
\cmidrule(r){1-5}
Class & Precision & Recall & F1-score & Support\\
\midrule
Benign & 90.61\% & 87.04\% & 88.79\% & 696827\\
Malware & 87.80\% & 91.18\% & 89.46\% & 424420 \\
\midrule
avg/total & 89.19\% & 89.13\% & 89.13\% & 1121247 \\
\midrule
False Positive Rate & 8.79\%\\
False Negative Rate & 12.96\%\\
\bottomrule
\end{tabular}
}
\end{table}

\section{Conclusion}

We have engineered a robust control flow graph-based vectorization scheme for exposing features which reveal semantically interesting constructs of .NET files.  The vectorization scheme is interpretable and glass-box by construction, which will facilitate scalable taxonomy operations in addition to high-accuracy classification as benign or malicious. 

The control flow-type graphs include both function call graphs, one for each file, and SDFG graphs, one for each function defined within the file. 
 Leveraging the combinatorial structure of these graphs results in a rich feature space, within which even a simple classifier can effectively distinguish between benign and malicious files.     

The vectorization scheme introduced here may be leveraged to train a standalone model or to augment the feature space of an existing model.  Although we limited our experiments to decompiled .NET, we see no obstruction to applying the PMIV concept to a wider class of graph-based file data, such as disassembly. 

Future work will involve the addition of new functions $f:\mathrm{Vert}(G)\longrightarrow \mb{R}$ for control flow-type graphs $G$ to the vectorization scheme, as well as the clustering of files and the functions of which they consist within both the codomain of the vectorization map and within the graph space $\Gamma$.  We will also explore the extent to which these functions and files can be parameterized through manifold learning in Euclidean as well as graph space.

\section*{Acknowledgment}

The authors would like to thank former colleague Brian Wallace for both deduplicating our .NET corpus and applying the decompiler at scale.  Without his efforts, this project would not have been possible.



%


\appendix

\subsection{Complete List of Functions on SDFGs}\label{FUNCSONSDFGS}
All functions are assumed to be zero on nodes for which the associated AST member is inconsistent with the function definition.  For example, NumPass2Call is trivial on all non-Call nodes. 
\begin{itemize}
\item $\mm{ExpectedType}$ \begin{equation}
       v,s\mapsto
        \begin{cases}
           1 & \text{if} \,\,\mm{type}(v)==s \\ \nonumber
            0 & \text{otherwise}
        \end{cases}
    \end{equation}
    
\item $\mm{CLRVariable}:v\mapsto  \eta(\mm{varType}(v))$
\item $\mm{BinaryOp}:v\mapsto  \eta(\mm{whichOpCode}(v)) $
\item $\mm{CtorCallctorType}:v\mapsto \eta(\mm{ctorType}(v))$
\item $\mm{FieldReference}:v\mapsto \eta(\mm{fieldName}(v))$
\item $\mm{CLRLiteral}:v\mapsto \mm{value}(v)$
\item $\mm{CallfnName}:v\mapsto \eta(\mm{fnName}(v))$
\item $\mm{CLRArrayelemType}:v\mapsto \eta(\mm{elemType}(v))$
\item $\mm{FnPtrObjname}:v\mapsto \eta(\mm{name}(v))$
\item $\mm{TypeTesttestedType}:v\mapsto \eta(\mm{testedType}(v))$
\item $\mm{ClassRefname}:v\mapsto \eta(\mm{name}(v))$
\item $\mm{TypeCast}:v\mapsto \eta(\mm{castedType}(v))$
\item $\mm{CLRArraysize}:v\mapsto \eta(\mm{elemType}(v))$
\item $\mm{NumPass2Call}:v\mapsto \#(\mm{arguments}(v))$
\item $\mm{AddressOf}:v\mapsto \mm{expr}(v)$
\item $\mm{ThrowOpexpr}:v\mapsto \mm{expr}(v)$
\item $\mm{UnaryOpexpr}:v\mapsto \mm{expr}(v)$
\item $\mm{StoreLocallocalIdx}:v\mapsto \mm{localIdx}(v)$
\item $\mm{StoreLocalvalue}:v\mapsto \mm{value}(v)$
\item $\mm{Returnvalue}$\begin{equation}
        v\mapsto
        \begin{cases}
            \mm{value}(v) & \text{if} \,\,\mm{type}(\mm{value}(v))==\mm{float} \\ \nonumber
            0 & \text{if} \,\, \mm{type}(\mm{value}(v))==\mm{dict}
        \end{cases}
    \end{equation}
\end{itemize}
Note that value$(v)$, expr$(v)$ are floats and $\#\mm{arguments}(v),\mm{localIdx}(v)$ are ints.

\subsection{CLR AST Dictionary}\label{DICT}
The dictionary of terms relating to the CLR is as follows:


\includegraphics[scale=0.45,width=8cm]{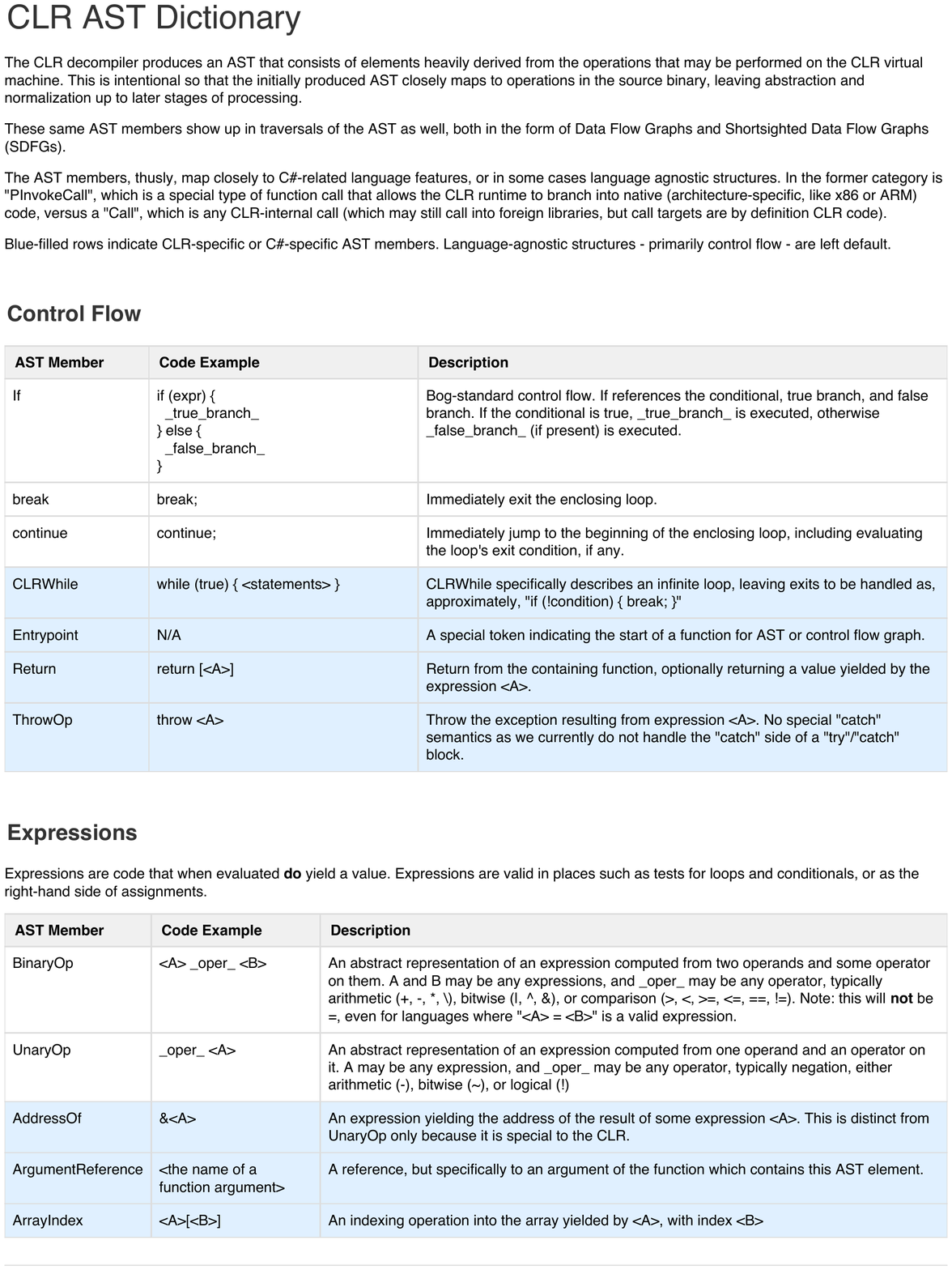}

\includegraphics[scale=0.45,width=8cm]{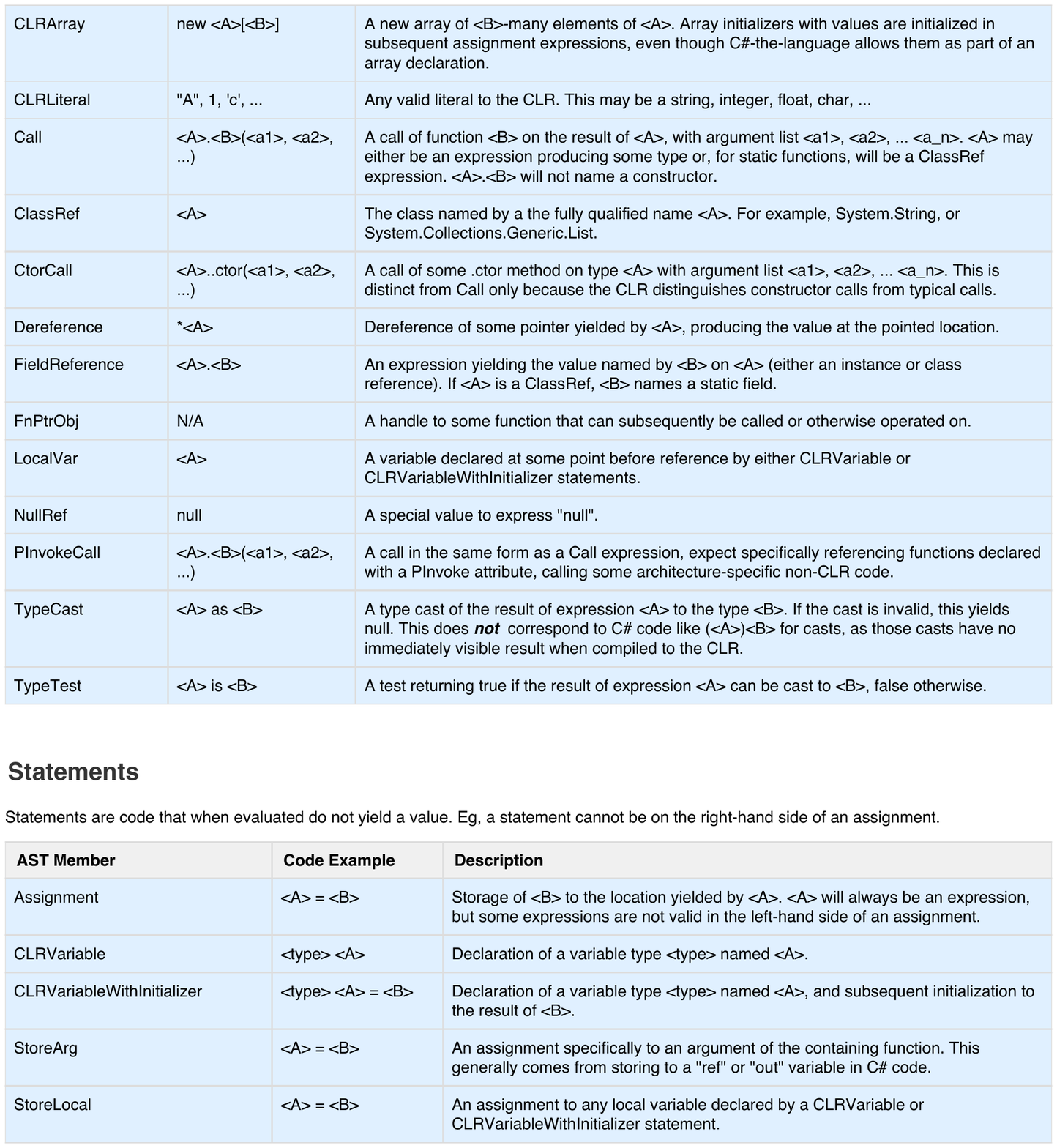}

%
%

\end{document}